\documentclass[runningheads]{llncs}
\usepackage{graphicx}
\usepackage{comment}
\usepackage{amsmath,amssymb} 
\usepackage{color}
\usepackage{epsfig}
\usepackage{subfigure}
\usepackage{multirow}
\usepackage{colortbl}
\usepackage{arydshln}
\usepackage{lipsum,mwe,cuted}
\usepackage{multirow}
\usepackage{float}
\usepackage{mathtools,xparse}

\begin{document}
\pagestyle{headings}
\mainmatter
\def\ECCVSubNumber{3495}  

\title{NormalGAN: Learning Detailed 3D Human from a Single RGB-D Image} 
\titlerunning{NormalGAN}
\author{Lizhen Wang\inst{1}\orcidID{0000-0002-6674-9327} \and
Xiaochen Zhao\inst{1}\orcidID{0000-0001-8976-7723} \and
Tao Yu\inst{1}\orcidID{0000-0002-3818-5069} \and
Songtao Wang\inst{1}\orcidID{0000-0003-2203-1572} \and
Yebin Liu\inst{1}\orcidID{0000-0003-3215-0225}}
\authorrunning{L. Wang et al.}
\institute{Tsinghua University, Beijing, China}
\maketitle

\begin{abstract}
We propose NormalGAN, a fast adversarial learning-based method to reconstruct the complete and detailed 3D human from a single RGB-D image. 
Given a single front-view RGB-D image, NormalGAN performs two steps: front-view RGB-D rectification and back-view RGB-D inference. 
The final model was then generated by simply combining the front-view and back-view RGB-D information. 
However, inferring back-view RGB-D image with high-quality geometric details and plausible texture is not trivial.
Our key observation is: Normal maps generally encode much more information of 3D surface details than RGB and depth images. Therefore, learning geometric details from normal maps is superior than other representations. 
In NormalGAN, an adversarial learning framework conditioned by normal maps is introduced, which is used to not only improve the front-view depth denoising performance, but also infer the back-view depth image with surprisingly geometric details. 
Moreover, for texture recovery, we remove shading information from the front-view RGB image based on the refined normal map, which further improves the quality of the back-view color inference. 
Results and experiments on both testing data set and real captured data demonstrate the superior performance of our approach. 
Given a consumer RGB-D sensor, NormalGAN can generate the complete and detailed 3D human reconstruction results in 20 fps, which further enables convenient interactive experiences in telepresence, AR/VR and gaming scenarios. 
   
\end{abstract}

\section{Introduction}
\label{sec:intro}

Reconstructing 3D human from RGB or RGB-D images is a popular research topic with a long history in computer vision and computer graphics for the reason of its wide applications in image/video editing, movie industries, VR/AR content creation, etc. 

The recent trend of using a single depth camera has enabled many applications especially in 3D dynamic scenes and objects reconstruction\cite{newcombe2015dynamic,doublefusion,yu2017BodyFusion}. 
Using depth images, the depth ambiguity of monocular inputs is resolved as it is straight-forward to calculate the 3D coordinate of each pixel. 
However, in order to obtain a complete model, traditional reconstruction approaches~\cite{GuoTOG2017,yu2017BodyFusion} using depth cameras usually require multi-frame information with tedious capture procedures. 
Moreover, 3D models of the subjects have to maintain a fixed geometric topology in the whole sequence in current fusion-based methods, which limits the ability to handle topology changes during capture.
Finally, the tracking-based methods~\cite{doublefusion} need to fuse multi-frame RGB-D information which will result in over-smoothed geometry and texture. 

On the other end of the spectrum, recent progress has shown impressive capability of using learning-based techniques to reconstruct 3D human from a single RGB image~\cite{varol2018bodynet:,deephuman,moulding,pifu}.
However, current learning-based methods pay more attention on learning shapes but not surface details of the subject, which makes them struggle from recovering 3D models with complete geometric details when comparing with multi-view or multi-image-based methods(~\cite{Dou_Fusion4D_2017,newcombe2015dynamic}).
Furthermore, due to the inherent depth ambiguity of a single RGB image, the generated 3D model can not describe the real world scale of the subject. 
Finally, due to the heavy network and the complicated 3D representation used in current works, it is hard for current learning-based methods to achieve real-time reconstruction performance. 

\begin{figure}[t]
	\begin{center}
		\includegraphics[width=0.9\linewidth]{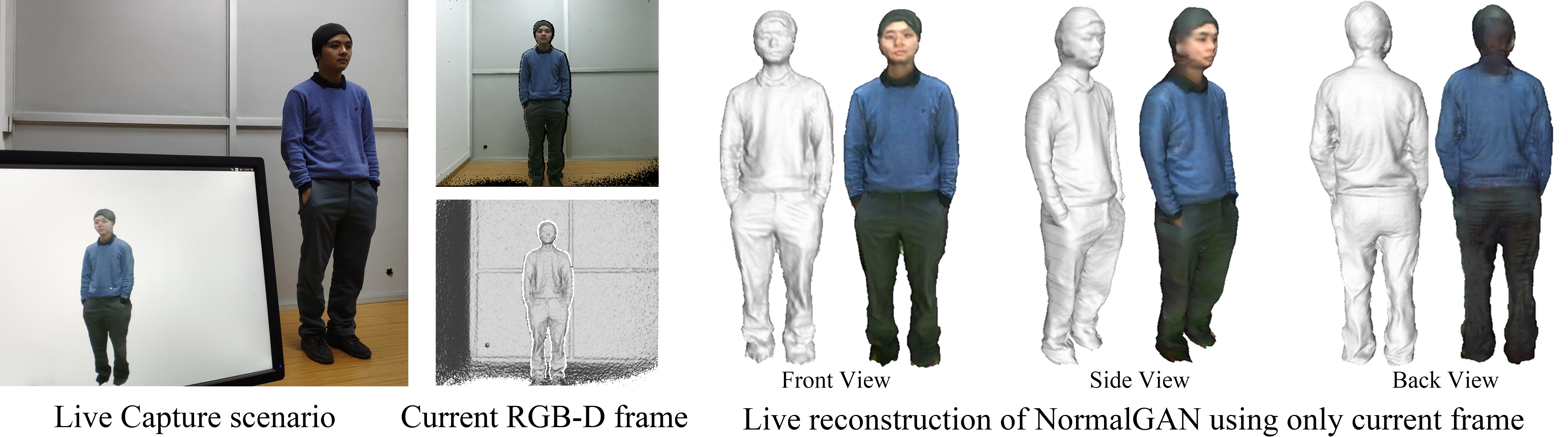}
	\end{center}
\caption{System setup and live reconstruction result of the current frame. We use a single Kinect V2 sensor for data capture.
}
\label{fig:cover}
\end{figure}

The most related work to our method is Moulding Humans~\cite{moulding}, which uses a single front-view RGB image to infer front-and-back-view-depth images. 
Although moulding human generates plausible 3D human reconstruction results, the final output models are over smoothed, lacking geometric details even in the visible regions. Moreover, texture recovery is not considered in Moulding Humans, as well as the real world size of the subject. 
To overcome the challenges in Moulding Humans, we incorporate depth information into our method, and use front-view RGB-D image as input to infer the back-view RGB-D image. Although the incorporation of the front-view depth image eliminates the depth ambiguities and provides real world scales, using a single front-view RGB-D image to infer a complete and detailed 3D human model remains challenging. The reasons include: 
1) RGB-D images captured by consumer RGB-D sensors always contain severe depth noise, which makes it extremely difficult to extract sharp geometric details while smoothing severe noise, and texture copy artifacts are inevitable even using the most recent learning based depth denoising methods~\cite{yan2018DDRNet}; 
2) Although GAN-based methods have shown the capability of inferring the invisible area, they can only generate a reasonably over-smoothed results without any details. Moreover,
even with a noise-free depth image, it remains challenging to learn geometric details from depth values directly. Compared with the depth variance on the global body shape, the depth variance around local geometric details is too subtle to ''learn'' for the deep neural networks. 
3) The performance of inferring back-view RGB from front-view RGB is unstable due to the various shading effects coupled in the RGB images. 

To tackle the above problems, we propose NormalGAN, a novel adversarial framework conditioned by normal maps. 
The proposed method can generate high-quality geometry with complete details given a single front-view RGB-D image. Different from previous methods that directly constrain each pixel on the depth maps, NormalGAN also focuses on constraining the adjacent relations of each pixel. Specifically, we use normal maps, in which pixel values change drastically around geometric details, instead of depth images to condition a GAN, and enable the generation of more reasonable and realistic geometric details for both front-view depth denoising and back-view depth inference. 
Moreover, an RGB-D rectification module based on orthographic projection and neural networks is proposed and significantly improves the reconstruction quality around body boundaries and extremities. The rectified RGB-D images achieve more regular sampling of the human body than the raw perspective projected RGB-D images from sensors. 
Finally, to further improve the performance of back-view RGB inference, we decompose the various shading information from the front-view RGB images to factor out the misleading impacts of shading variations. 

Our technical contributions can be concluded as: 1) A novel method to recover complete geometric details from both the front-view depth (denoising) and the back-view depth (inference) using an adversarial learning framework conditioned by normal maps. 2) An RGB-D rectification module based on orthographic projection and neural networks to enable regular RGB-D sampling of the human body. 3) A method for back-view RGB inference in the intrinsic domain, which further improves the texture inference quality in the invisible area.


\section{Related Work}\label{sec:related_work}

\textbf{3D Human from RGBD sequences.}
For 3D human body modeling, the most intuitive method is to get enough RGB or depth observations from different viewpoints first, and then fuse them together. 
KinectFusion~\cite{Izadi:2011} is the pioneer work in this direction which combines rigid alignment with volumetric depth fusion in an incremental manner. 
Subsequent works~\cite{tong2012scanning,cui2013kinectavatar,Zeng_2013_CVPR} combined KinectFusion with non-rigid bundling for more accurate reconstruction. 
Moreover, to handle dynamic scenes, DynamicFusion~\cite{newcombe2015dynamic} contributes the first method for real-time non-rigid volumetric fusion. 
Follow up works keep improving the performance of DynamicFusion by incorporating non-rigid deformation constraints~\cite{innmann2016volume,slavcheva2017cvpr,Slavcheva_2018_CVPR}, articulated motion prior~\cite{yu2017BodyFusion,chao2018ArticulatedFusion}, appearance information~\cite{GuoTOG2017}, and parametric body shape prior~\cite{doublefusion}. 
Due to the incremental fusion strategy used in current fusion-based methods, subjects are required to maintain a fixed geometry topology in the whole sequence, which limits the usage of such methods for more general scenarios.

\noindent\textbf{Human Body Reconstruction from a single image.}
Rencent studies focusing on reconstructing a 3D human model from a single image adopt various representations including the parametric human model, depth map, silhouette and implicit function. Primitive approaches require strong priors of parametric human models, which are widely used for human shape and pose estimation~\cite{Guan:ICCV:2009,SMPL:2015,SMPL-X:2019,rhodin2016general}. BodyNet~\cite{varol2018bodynet:} first shows the capability of reconstructing a 3D human from only a single image employing SMPL~\cite{SMPL:2015} as a constraint.  Follow up approaches~\cite{kanazawa2018end-to-end,pavlakos2018learning} improve the robustness of parameters estimation for challenging images. DeepHuman~\cite{deephuman} employs SMPL~\cite{SMPL:2015} as a volumetric initialization and achieves stable results for complex poses. The geometric details like clothes folds, hair and facial features are mostly ignored.
Recently, Gabeur et al.~\cite{moulding} proposed a novel moulding representation, which divides the whole person into two parts. Employing two depth maps, they transfer the complicated 3D problem to an image-to-image translation problem, which significantly improves efficiency.
SiCloPe~\cite{natsume2019siclope:} generates fully textured 3D meshes using a silhouette-based representation.
Saito et al.~\cite{pifu} presents PIFu, which recovers high-resolution 3D textured surfaces. PIFu achieves state-of-the-art results by encoding the whole image into a feature map, from which the implicit function infers a textured 3D model. Besides, FACSIMILE~\cite{smith2019facsimile} reconstructs detailed 3D naked body by directly constraining normal maps, which helps to recover the geometric details of naked bodies. 
However, the inherent depth ambiguities of 2D-to-3D estimation from RGB images is inevitable. 

\noindent\textbf{Depth Denoising.}
Depth images are widely used in various applications, but consumer depth cameras suffer from heavy noise due to complex lighting conditions, the interference of light and so on. As RGB images usually have higher quality than depth images, many methods focus on enhancing depth images with the support of RGB images. Some methods leverage RGB images by investigating lighting conditions. Shape-from-shading techniques can extract geometric details from RGB images~\cite{petrov1993on,zhang1999shape-from-shading:}. 
Recent progress has shown that shading information helps to recover geometric details even in uncontrolled lighting conditions, with a depth camera~\cite{han2013high,yu2013shading-based}, or multi-view cameras~\cite{wu2011,wu2013}. However, recovering geometric details is still very challenging for traditional schemes due to the unknown reflectance. More recently, Yan et al.~\cite{yan2018DDRNet} proposes DDRNet to handle this problem with a learning-based framework, which can effectively remove the noise while recovering geometric details in a data-driven way. Afterwards, Sterzentsenko et al.~\cite{sterzentsenko2019self-supervised} improves the denoising qualities with multi-view RGB-D images. Despite the efforts, extracting accurate geometric information from RGB images is still a challenge. Our approach tackles this problem by constraining the normal maps in an adversarial network architecture.

\section{Overveiw}\label{sec:overview}

\begin{figure}[t]
\centering
\includegraphics[width=0.9\linewidth]{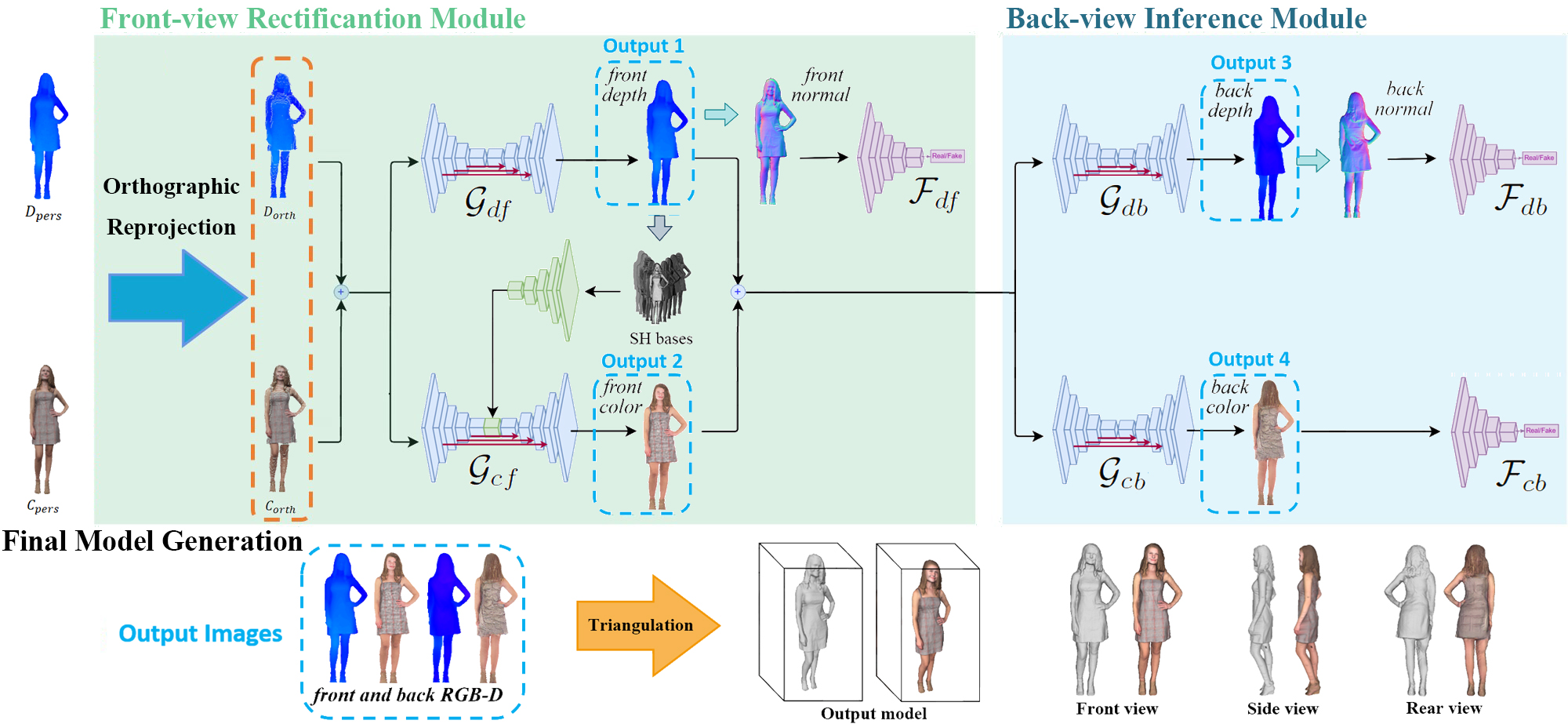}
\caption{The overview of our approach. }
\label{fig:pipe}
\end{figure}

The overview of our method is shown in Fig.\ref{fig:pipe}.
Our method takes a front-view RGB-D image (including a color image $C_{pers}$ and a depth image $D_{pers}$) as input, where the subscript $_{pers}$ indicates the input RGB-D image was captured using the perspective projection model. 
We then perform front-view rectification and back-view inference sequentially to get the rectified front-view RGB-D image (output1 and output2 in Fig.\ref{fig:pipe}) and the inferred back-view RGB-D image (output3 and output4 in Fig.\ref{fig:pipe}). 

For the front-view rectification, 
in order to avoid the artifacts caused by the irregular sampling property of the perspective projection, we first transform the RGB-D image into a colored 3D point cloud, and then render a new RGB-D image using the orthographic projection, which brings a pair of incomplete images, $C_{orth}$ and $D_{orth}$.
To further refine $D_{orth}$, we use a UNet $\mathcal{G}_{df}$ to inpaint the missing areas after orthographic re-projection, and, more importantly, to remove the severe depth noise. 
Note that it is difficult to prevent the traditional CNN-based depth denoising networks from over-smoothing the geometric details on the depth map. So we incorporate a discriminator $\mathcal{F}_{df}$, which is conditioned by normal maps, to enforce the adjacency relation of each point on the refined depth map, and finally guides $\mathcal{G}_{df}$ to generate the refined front-view depth (output1) with high-quality geometric details. 
Finally, to refine $C_{orth}$, we use the network $\mathcal{G}_{cf}$ and the shading information calculated from the refined depth maps~\cite{wu2013} for color inpainting and shading decomposition, and finally get the refined (intrinsic) front-view color image (output2). 

For back-view RGB-D inference, we use two CNNs $\mathcal{G}_{db}$ and $\mathcal{G}_{cb}$ combined with two discriminators ($\mathcal{F}_{db}$ and $\mathcal{F}_{cb}$) to infer the back-view depth and color image (output3 and output 4 in Fig.\ref{fig:pipe}), respectively.
The discriminator $\mathcal{F}_{db}$ is also conditioned by normal maps as in $\mathcal{F}_{df}$, which helps to generate reasonable and realistic geometric details on the back-view depth image. 
Moreover, the discriminator $\mathcal{F}_{cb}$ is conditioned by ground truth color, which enforces more plausible back-view color generation. 

After getting the front-view and back-view RGB-D image pair (output 1,2,3 and 4 in Fig.\ref{fig:pipe}) from our networks, we directly triangulate the resulting 3D points into a complete mesh with vertex color for final model generation (as shown in the bottom of Fig.\ref{fig:pipe}). Note that the final model is relighting-ready benefiting from the intrinsic-domain-color-inference.

\section{Method}
\label{cha::method}

\subsection{Dataset Generation}
\label{sec:datasets}
To generate our training and testing dataset, we use 1000 static 
3D scans of clothed people purchased from twindom\footnote{https://web.twindom.com/}. 
800 of them are used in train-dataset and the other 200 are used to test. While rendering, 800 3D scans are randomly rotated in the range between $-30^{\circ}$ and $30^{\circ}$ to enrich our training dataset. 
To obtain the supervised data of our networks, we first collect a dataset of RGB-D images $(C_{pers}, D_{pers})$ in the perspective view, together with corresponding ground-truth RGB-D images $(C_{gt}, D_{gt})$ in the orthographic view. All images are rendered with the resolution of $424\times{512}$, which is the same as that of depth maps captured by Kinect v2 sensors. The training code and pre-trained models will be public available. And please note that we cannot release the full training dataset due to business constraints.

\begin{figure}[t]
	\begin{center}
		\includegraphics[width=0.9\linewidth]{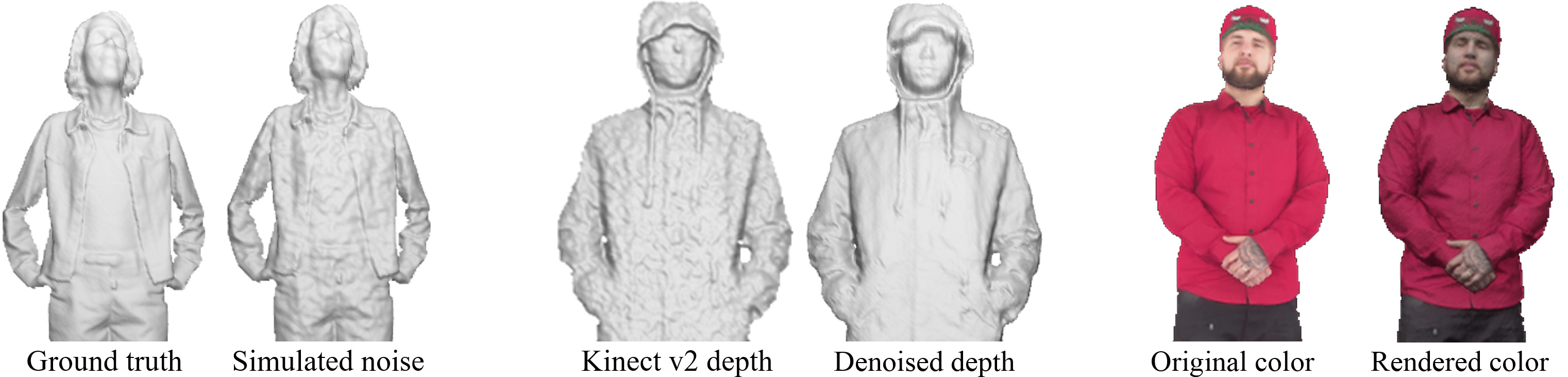}
	\end{center}
\caption{Left: we add simulated noise to the ground truth depth map. Middle: the networks trained on our simulated depth images show stability in denoising Kinect v2 depth images. Right: We render the models with random illuminations.}
\label{fig:noise}
\end{figure}

In order to make our networks applicable to real data, data augmentation is necessary. Firstly, we simulate the typical noise distribution of Kinect v2 depth cameras (see the left images in Fig.\ref{fig:noise}). Noticing that real noise distribution has local correlation rather than pixel-wise independence, we apply multiple 2D Gauss kernels on $D_{pers}$. The standard deviation of Gauss kernels are set as constant and the position of Gauss kernels are uniformly distributed over the image. To simulate the different noise levels of Kinect depth maps, we calculate the amplitude of the above Gauss kernel with the formulation proposed by P{\'e}ter et al.~\cite{fankhauser2015kinect}. Our networks trained on the simulated depth images show stability in Kinect v2 depth images (see the middle images in Fig.\ref{fig:noise}). Secondly, we render the models with random illuminations to get the more realistic color images, an example can be see in Fig.\ref{fig:noise}. Finally, we get a training set with 5,600 groups of images and a testing set with 1,400 groups of images in total.

\subsection{Front-view RGB-D Rectification}
\label{sec:front}
Our method adopts a double-sided representation for human bodies similar to Moulding Humans~\cite{moulding}, which facilitates the subsequent processing and is able to speed up with lightweight networks. The representation is based on the assumption of dividing a person into a visible front-view map and an invisible back-view map. However, in the perspective-view images, the visible part is actually smaller than a ``half'' of a person, which will cause the irregular sampling property. In order to solve the above problem, we first transform the perspective-view RGB-D images ($C_{pers}$ and $D_{pers}$) into 3D colored point clouds, then re-project them into the orthographic view. And the missing areas in the generated orthographic-view RGB-D images ($C_{orth}$ and $D_{orth}$) will be filled by the following networks. 


To get the complete and denoised front-view depth images, we use a generator $\mathcal{G}_{df}$. The basic structure of $\mathcal{G}_{df}$ is UNet~\cite{ronneberger2015u-net:}, which is commonly used for image-to-image translation tasks. However, it is difficult for a simple UNet to denoise the depth images while preserving the geometric details. So we incorporate a discriminator $\mathcal{F}_{df}$ and use the GAN architecture to enhance the denoising ablity of $\mathcal{G}_{df}$. As the shading information of the input color images indicates the geometric information, we take the color images $C_{orth}$ as input to support the refinement of $D_{orth}$. However, while training $\mathcal{G}_{df}$ with only depth maps, the output depth images usually become over-smooth and complex textures usually lead to wrong geometric details as the network can not distinguish the geometric details from the texture changes. Even if we use the GAN structure, there is no noticeable change. This is because the depth variance around local geometric details is too subtle compared with the depth variance on the global body, which is also demonstrated by FACSIMILE~\cite{smith2019facsimile}. However, FACSIMILE mainly focuses on A-pose naked body recovery by directly constraining normal maps with L1 loss, and this can not be used to handle the clothed bodies directly without GAN loss. Therefore, we apply GAN on normal maps to recover geometric details of the clothed bodies. The normal map can be described as
\begin{equation}
\label{equ:normal}
\mathbf{N}_{i} = Norm(\sum\limits_{j,k} {Norm((\mathbf{P}_{j}-\mathbf{P}_{i}) \times (\mathbf{P}_{k}-\mathbf{P}_{i}))})
\end{equation}
where $\mathbf{N}_{i}$ is the normal vector of point $i$, and $\mathbf{P}_{i}$ is the 3D coordinate of point $i$. The function $Norm(\cdot)$ normalizes the input vector. Point $j$ is in the clockwise direction of point $k$ relative to point $i$ and both of them belong to the neighborhood of point $i$. 

Obviously, normal maps have noticeable changes around the geometric details compared with depth maps. Additionally, normal maps show the adjacency relations of each point on the depth map, which are important for generating the geometric details.
However, note that it is not straightforward to use the normal information as the drastically changing normal values may cause numerical crash during the training process. Therefore, we add additional loss terms for the GAN structure (See Sec.\ref{sec:loss}) and pretrain $\mathcal{G}_{df}$ before joint training. Finally, $\mathcal{F}_{df}$ conditioned by normal maps guides $\mathcal{G}_{df}$ to generate denoised depth maps with geometric details preserved. 

Then, to get high-quality textures of human bodies, we use another UNet $\mathcal{G}_{cf}$ to remove the shading from the input $C_{orth}$. 
For the color images captured in real-world scenes, various shading effects are coupled in the RGB images, which have great influence on inferring the invisible textures. In order to remove the shading effects, we encode the shading information calculated by the spherical harmonics function~\cite{wu2013} from the refined depth images and concatenate it with the tensor of $\mathcal{G}_{cf}$ (See the green network in Fig.\ref{fig:pipe}). Finally, the trained $\mathcal{G}_{cf}$ shows good capability on shading removal and brings the refined (intrinsic) front-view color images.

\subsection{Back-view RGB-D Infernence}

\begin{figure}[t]
	\begin{center}
		\includegraphics[width=0.8\linewidth]{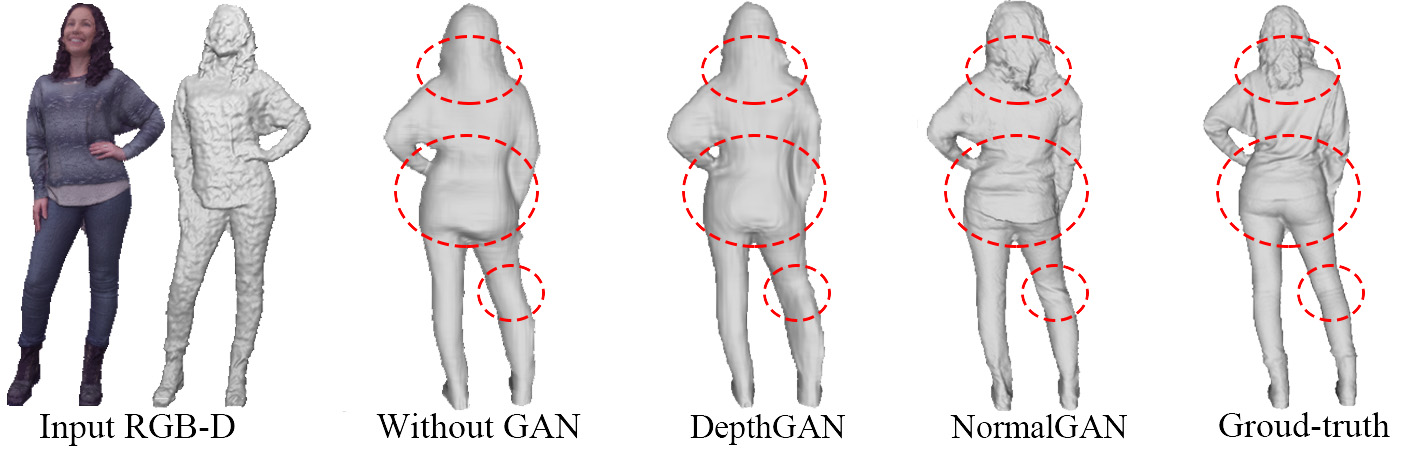}
	\end{center}
\caption{Results of our method without GAN, DepthGAN (GAN conditioned by depth maps) and NormalGAN.}
\label{fig:normalGAN}
\end{figure}

Obtaining the rectified front RGB-D images, we need to infer the invisible back-view RGB-D images with reasonable details. As it is difficult for traditional CNN-based methods to infer the details, we also use the GAN structure to enhance the inference of our generators. Specifically, for the bakc-view depth inference networks (the generator $\mathcal{G}_{db}$ and the discriminator $\mathcal{F}_{db}$), we still use normal maps to condition the $\mathcal{F}_{db}$ which helps to generate reasonable geometric details. As shown in Fig.\ref{fig:normalGAN}, our GAN conditioned by normal maps shows great superiority in inferring the geometric details on the back side. Furthermore, the different details of NormalGAN and ground truth eliminate the possibility of over-fitting. For the back-view color inference networks $\mathcal{G}_{db}$ and $\mathcal{F}_{cb}$, we condition the GAN by the ground truth color images, which helps to infer more plausible textures. Moreover, the shading removal of front-view color images shows great influence on the back-view color inference (See Fig.\ref{fig:texture}). Additionally, due to the orthographic re-projection, the front-view images and the back-view images share the same silhouette, which makes the inference simple but reasonable. Finally, plausible and realistic inferred details are generated in the back-view RGB-D images by our networks.

In the end, after getting the colored 3D point clouds from the front-view and the back-view RGB-D images, we triangulate the points by directly connecting the adjacent points and get the 3D meshes of the human bodies. However, there still remains a gap between front-view and back-view depth boundaries for mesh reconstruction. To fill this gap, we first interpolate three points between each boundary pairs (the same pixel on the boundary of front-view and back-view depth images), and then stitch the adjacent points.

\subsection{Loss Functions}
\label{sec:loss}

To ensure the convergence of the above networks, we put independent constraints on each network. First of all, L1-norm term is necessary for typical supervised training process. The L1 loss of the output $\mathbf{X}_{pred}$ and the ground-truth $\mathbf{X}_{gt}$ can be expressed as
\begin{equation}
\label{equ:l1 loss}
\mathcal{L}_{L1}(\mathbf{X}) = \lVert\mathbf{X}_{pred} - \mathbf{X}_{gt}\rVert_{1}
\end{equation}

To generate more sharp details for color images, we also use the perceptual loss~\cite{johnson2016perceptual}. Here, we use the pretrained VGG19~\cite{simonyan2014very} to extract the features from the color images. Denoting the activations of the $i$th layer in the VGG network as $\phi_i$ and representing the weights of the $i$th loss term with $\lambda_i$, the perceptual loss can be expressed as
\begin{equation}
\label{equ:perc loss}
\mathcal{L}_{Perc}(\mathbf{X}) = \sum\limits_{i} {\lambda_i\lVert\phi_i(\mathbf{X}_{pred}) - \phi_i(\mathbf{X}_{gt})\rVert_{1}}
\end{equation}

For the adversarial parts, the typical GAN loss: 
\begin{equation}
\label{equ:ad loss}
\begin{aligned}
\mathcal{L}_{GAN}(\mathbf{X}) = &\mathbb{E}_{\mathbf{X}_{gt}}[log \mathcal{F}(\mathbf{X}_{gt})] +
\mathbb{E}_{\mathbf{X}_{pred}}[log (1-\mathcal{F}(\mathbf{X}_{pred}))] 
\end{aligned}
\end{equation}
is necessary, where the discriminator $\mathcal{F}$ tries to maximize the objective function to distinguish the inferred images from the ground-truth images. By contrast, the generator $\mathcal{G}$ aims to minimize the loss to make the inferred images similar to the ground-truth images. 

Trained only with the GAN loss, our GANs usually fail to converge due to the dramatic numerical changes caused by normal maps. So we introduce the feature matching loss inspired by pix2pixHD~\cite{wang2018high} to further constrain the discriminators. Specifically, we calculate the loss of multi-scale feature maps extracted from the layers of a discriminator. The feature matching loss can effectively stabilize the parameters of the discriminator. We denote the $k$th-layer feature map of the discriminator as $D_{[k]}$ and the total number of layers as $N$. The loss can be expressed as: 
\begin{equation}
\label{equ:fm loss}
\mathcal{L}_{FM}(\mathbf{X}) = \sum\limits_{k=2}^{T-1}  {\lVert D_{[k]}(\mathbf{X}_{pred}) - D_{[k]}(\mathbf{X}_{gt}) \rVert}_{1}
\end{equation}

Finally, the loss functions for each network are formulated as
\begin{equation}
\label{equ:gen loss}
\begin{aligned}
\mathcal{L}_{\mathcal{G}_{df}} = & \mathcal{L}_{L1}(D_{front}) + \mathcal{L}_{Perc}(N_{front}) +  
                       \mathcal{L}_{FM}(N_{front}) + \mathcal{L}_{GAN}(N_{front}) \\
\mathcal{L}_{\mathcal{G}_{cf}} = & \mathcal{L}_{L1}(C_{front}) + \mathcal{L}_{Perc}(C_{front}) \\
\mathcal{L}_{\mathcal{G}_{db}} = & \mathcal{L}_{L1}(D_{back}) + \mathcal{L}_{FM}(N_{back}) + \mathcal{L}_{GAN}(N_{back}) \\
\mathcal{L}_{\mathcal{G}_{cb}} = & \mathcal{L}_{L1}(C_{back}) + \mathcal{L}_{FM}(C_{back}) + \mathcal{L}_{GAN}(C_{back}) \\
\mathcal{L}_{\mathcal{F}_{df}} = &- \mathcal{L}_{GAN}(N_{front}) \\
\mathcal{L}_{\mathcal{F}_{db}} = &- \mathcal{L}_{GAN}(N_{back}) \\
\mathcal{L}_{\mathcal{F}_{cb}} = &- \mathcal{L}_{GAN}(C_{back})
\end{aligned}
\end{equation}

\section{Experiments}
\label{cha:experiments}

\begin{figure}[t]
	\begin{center}
		\includegraphics[width=0.7\linewidth]{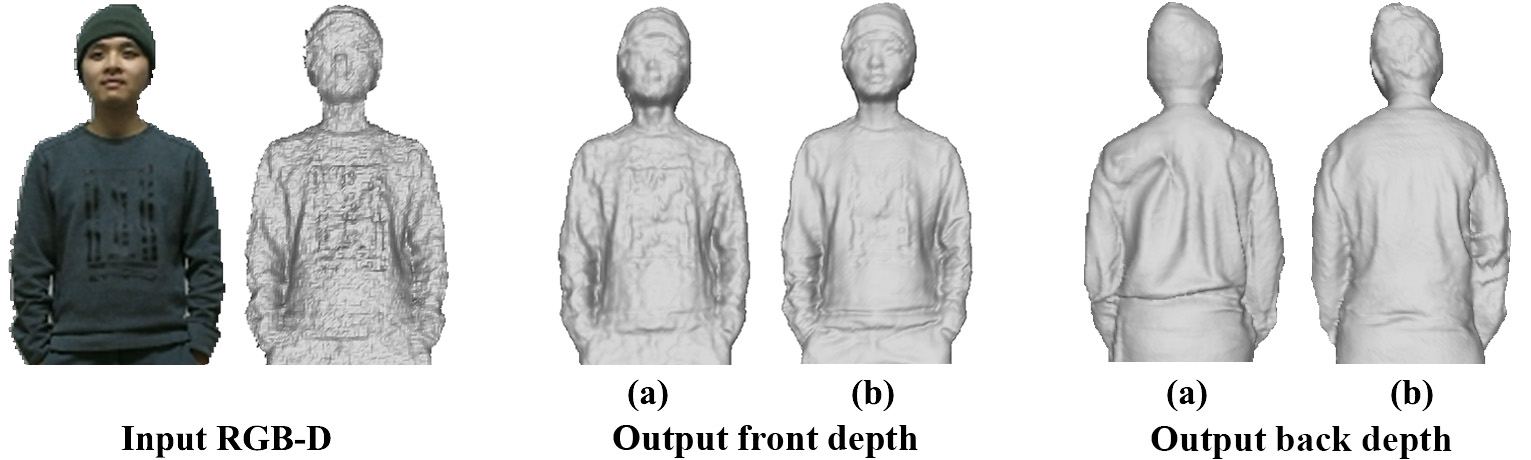}
	\end{center}
\caption{\textbf{(a)}: Results without using data augmentation. \textbf{(b)}: Results using data augmentation. The input RGB-D was captured by a Kinect v2 camera.}
\label{fig:dataaug}
\end{figure}

\subsection{Ablation Study}
\label{sec:ablation}
\textbf{The Validity of Data Augmentation.}
To demonstrate the validity of data augmentation in Sec.\ref{sec:datasets}, we train a reference model using the unprocessed dataset for comparison. 
As shown in Fig.\ref{fig:dataaug}, without our simulated noise, the reference network can not clearly remove the noise from the real-captured depth image and the back-view inference becomes unstable. In contrast, the data augmentation solves these problems and significantly improves our applicability for images captured in real scenes. 

\begin{table}[b]
\footnotesize
\begin{center}
    \begin{tabular}{l|c|c}
    \hline
    Error Metric                  & MAE       & RMSE   \\ \hline
    Input                         & 7.09     & 18.70  \\
    DDRNet~\cite{yan2018DDRNet}   & 5.00     &  7.49  \\
    Ours without GAN              & 4.75     &  7.38  \\
    Our DepthGAN                  & 5.02     &  7.43  \\
    Ours                          & \textbf{4.63}     & \textbf{7.18}  \\ \hline
    \end{tabular}
\caption{Quantitative comparisons of depth denoising on our testing set.}
\label{tab:rmse}
\end{center}
\end{table}

\begin{figure}[t]
	\begin{center}
		\includegraphics[width=0.8\linewidth]{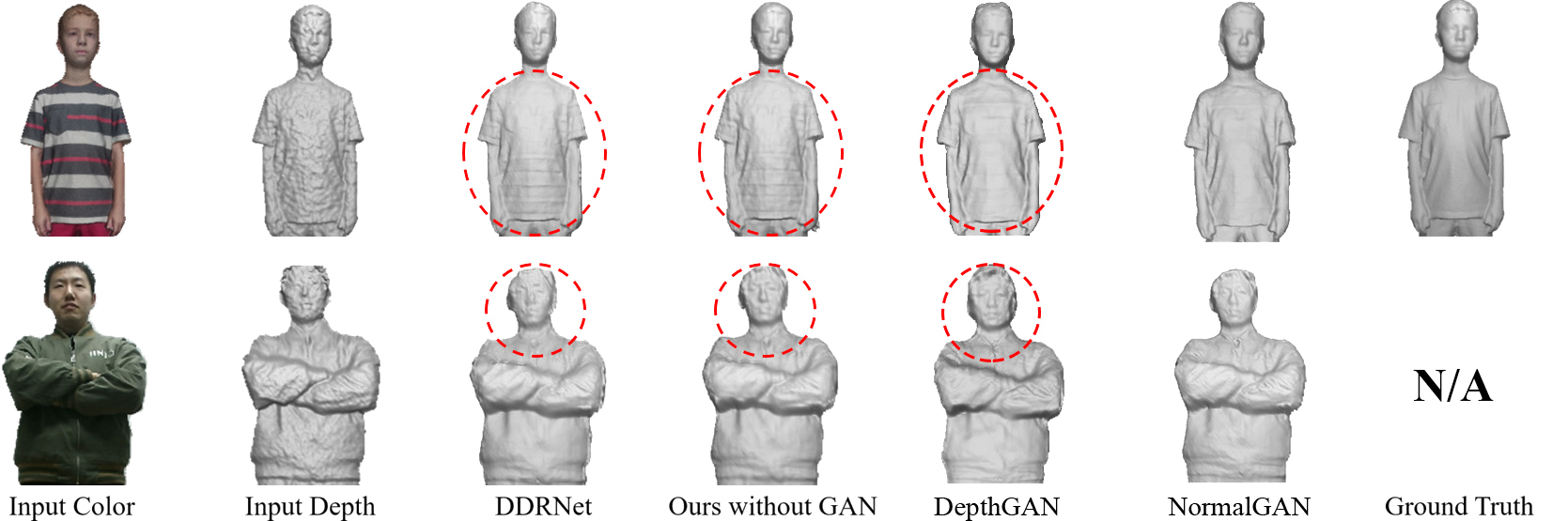}
	\end{center}
\caption{
Validation of depth denoising on both testing data and real captured data. 
}
\label{fig:ddrnet}
\end{figure}

\textbf{The Capability of Depth Denoising and Details Preserving.}
Using normal maps with the GAN structure, the capability of our networks to generate geometric details is significantly improved. For comparison, we train the state-of-the-art depth denoising network, DDRNet~\cite{yan2018DDRNet}, on our dataset. As shown in Fig.\ref{fig:ddrnet}, all the networks achieve effective results in depth denoising. However, from the first row in Fig.\ref{fig:ddrnet}, we can see that DDRNet, our network without GAN, and our GAN conditioned by depth maps (DepthGAN) all suffer from texture-copy artifacts (copying texture changes as geometric details). Besides, the second row shows the ability of our method to preserve more accurate details like facial expressions and wrinkles on the cloth. Finally, Tab.\ref{tab:rmse} gives the quantitative comparison on depth denoising accuracy, which further validates the effectiveness of our proposed method. 

\begin{figure}[b]
	\begin{center}
		\includegraphics[width=0.9\linewidth]{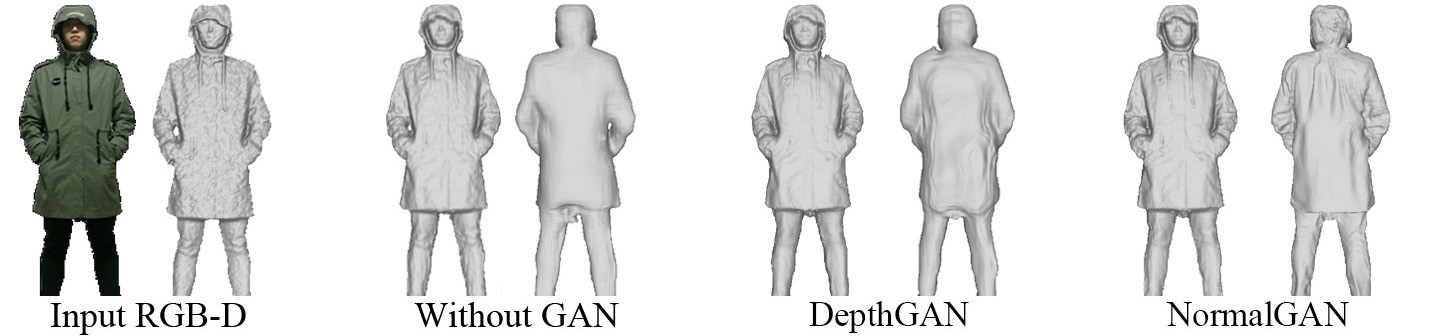}
	\end{center}
\caption{Comparsion of our method without GAN, DepthGAN (our GAN input with depth images) and NormalGAN using real captured data.}
\label{fig:GAN}
\end{figure}

\textbf{The Superiority of Our Method in Back-view Depth Inference.}
GAN with normal maps has significantly greater ability to infer the geometry of invisible areas. As we can see in Fig.\ref{fig:normalGAN}, the network trained without GAN and even DepthGAN can only generate over-smoothed results for reducing the L1 losses. However, using normal maps, the proposed NormalGAN can produce realistic and reasonable results despite the differences between the inferred details and the corresponding ground truth. Our method with normal maps also achieves more detailed results for images captured in real scenes as shown in Fig.\ref{fig:GAN}.

\begin{figure}
	\begin{center}
		\includegraphics[width=0.85\linewidth]{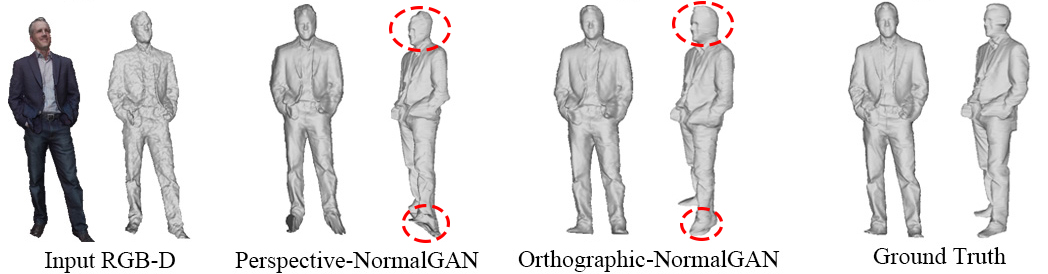}
	\end{center}
\caption{Comparison of NormalGAN trained on the perspective-view dataset and the orthographic-view dataset.}
\label{fig:orth}
\end{figure}

\begin{figure}
	\begin{center}
		\includegraphics[width=0.85\linewidth]{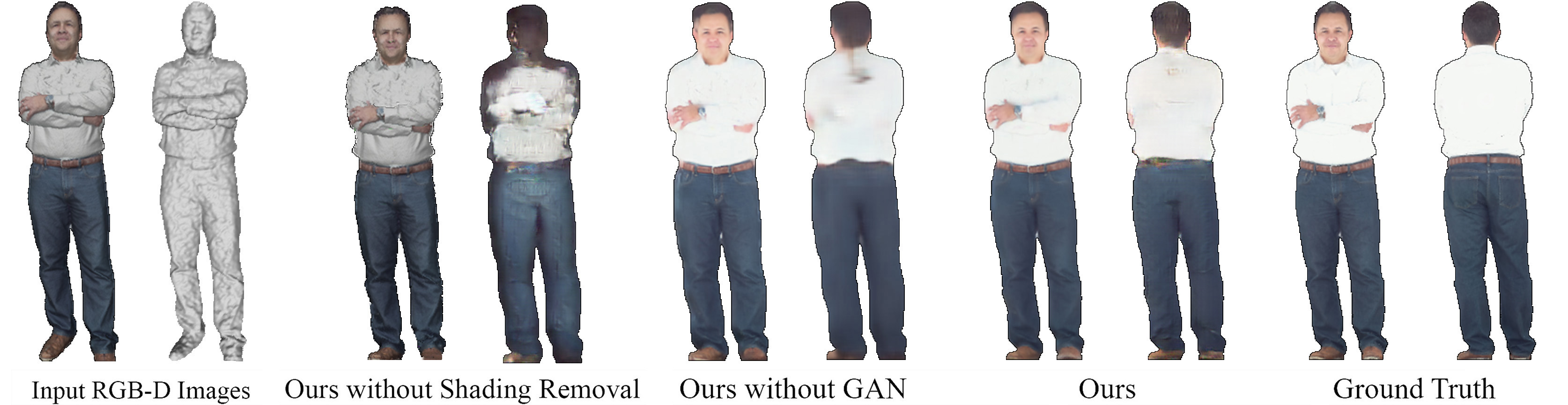}
	\end{center}
\caption{Texture results of our method with/without GAN.}
\label{fig:texture}
\end{figure}

\textbf{Orthographic Reprojection}
The orthographic projection helps our method to reconstruct more plausible results around the boundary areas and body extremities. 
As we can see in Fig.\ref{fig:orth}, the performance of networks trained under the perspective view decreases  around body extremities like hair and feet.

\textbf{The Effectiveness of Shading Removal.}
To remove shading from input color images, we encode the geometric information using the SH (spherical harmonics) bases. This module improves the applicability for real captured images. As shown in Fig.\ref{fig:texture}, the network without shading removal fails to handle the complicated shading caused by cloth wrinkles. Additionally, the shading removal helps to enhance the back-view color inference.

More detailed experiments are presented in the supplementary materials.

\begin{figure}[b]
\centering
\subfigure[Quantitative Evaluation]{
\label{fig:quant} 
\includegraphics[width=0.3\linewidth]{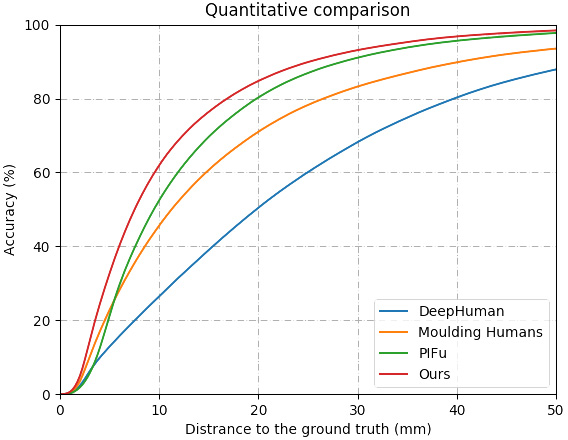}}
\subfigure[Error Maps]{
\label{fig:errormaps} 
\includegraphics[width=0.63\linewidth]{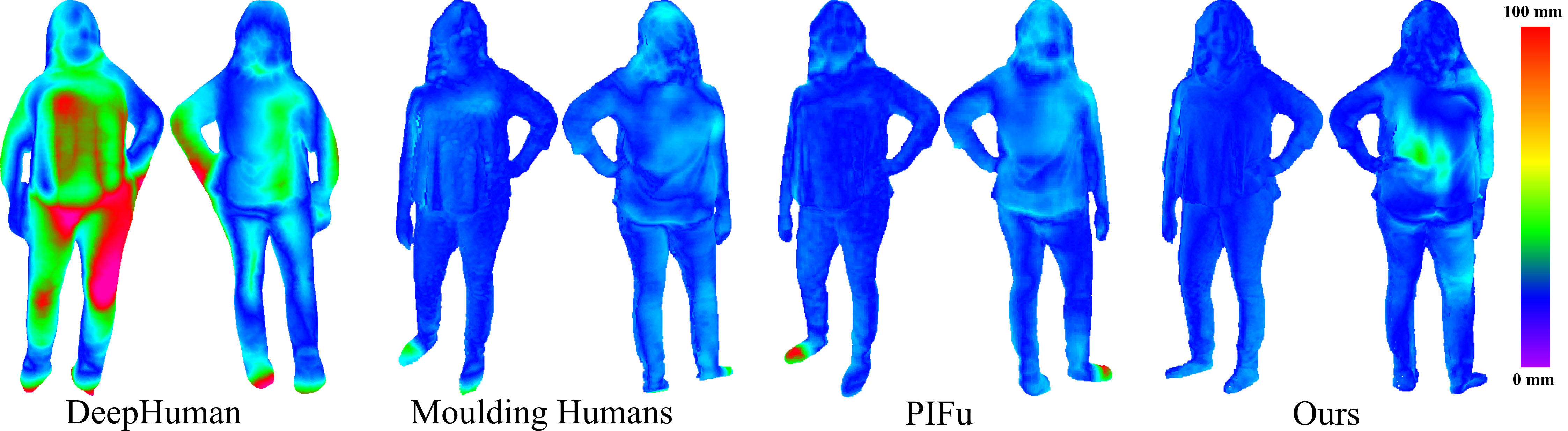}}
\caption{(a) Quantitative evaluation of our method and the retrained single-image methods. (b) Error maps of our method and the retrained single-image methods on one of the testing models. Errors above 100 mm are shown in red.}
\label{fig:evaluations} 
\end{figure}

\begin{figure}[h]
	\begin{center}
		\includegraphics[width=0.8\linewidth]{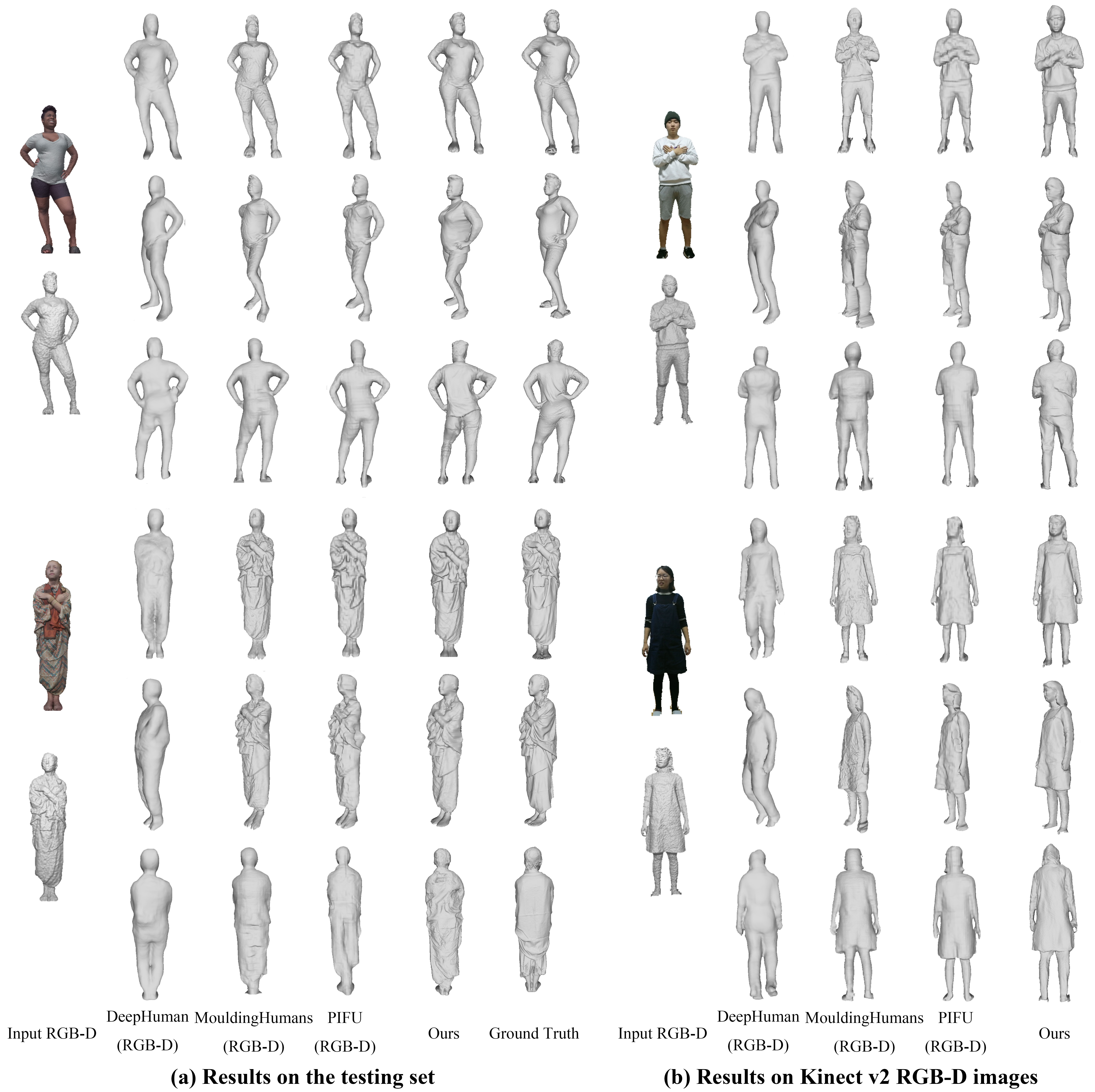}
	\end{center}
\caption{We evaluate our approach with the retrained DeepHuman (RGBD)~\cite{deephuman}, Moulding Humans (RGBD)~\cite{moulding} and PIFu (RGBD)~\cite{pifu} on testing dataset (the first three columns) and images captured by a Kinect v2 depth camera (the last two columns).}
\label{fig:all}
\end{figure}

\subsection{Comparison}
\label{sec:comparison}
\textbf{Single-image methods.}
As previous single-image methods to reconstruct the human bodies all take color images as their input, we retrain the methods on our dataset with RGB-D images. Specifically, for DeepHuman~\cite{deephuman}, we use depth images to fit more accurate SMPL models as its initialization model. For PIFU~\cite{pifu} and Moulding Humans~\cite{moulding}, we retrain the networks with our dataset by concatenating the RGB image and the depth image into a 4-channel RGB-D image as the new input for Moulding Humans and PIFu, which is exactly the same input as ours. Besides, the results of the above methods using RGB only are also presented in our supplementary materials.

As shown in Fig.\ref{fig:all}, our approach generate results with more geometric details, which indicates our capability to reconstruct details human models with the support of normal maps. 
The comparisons on images captured by a Kienct v2 depth camera demonstrate our capability to reconstruct the persons in real scenes. Then, we select 12 models in common poses from our testing set which all methods can generate correct 3D models. As shown in Fig.\ref{fig:quant}, our method also achieve the best result in the quantitative evaluation. Besides, the error maps in Fig.\ref{fig:errormaps} also demonstrate the superiority of our method.

\noindent\textbf{Multi-frame RBG-D methods.} We extensively evaluate our method with the volumetric fusion methods DoubleFusion~\cite{doublefusion} and Guo et al.~\cite{GuoTOG2017}. The results are presented in the supplementary materials.

\subsection{Training Details and Runtime Performance}
\label{sec:speed}
During the training procedure, we first train the four generators separately to accelerate the convergence. Then, after the convergence of each network, we train them together with the discriminators. The input images are cropped to the resolution of $424\times424$. All networks converge after 60 epochs of our training set. The training procedure takes about 8 hours on four GTX 1080Ti graphics cards. As the discriminators will not calculate in the testing procedure, our testing procedure occupies 3.5-GB GPU memory. Generating a 3D human model takes 50 ms on average using a RTX Titan.

\section{Conclusion}
\label{sec:discussion}

In this paper, we have presented NormalGAN, an adversarial learning-based method to reconstruct a complete and detailed 3D human from a single RGB-D image. 
By \textit{learning details from normal maps}, our method can not only achieve high-quality depth denoising performance, but also infer the back-view depth image with much more realistic geometric details. 
Moreover, the proposed RGB-D rectification module based on orthographic projection is effective for generating reasonable boundaries, and can be used in other related topics. 
Finally, with the support of shading removal in the front-view RGB images, we can improve the quality of back-view color inference. 
A live capture system was implemented and can robustly run at 20 fps, which further demonstrates the effectiveness of our method. 

Similar to Moulding Humans~\cite{moulding}, NormalGAN is based on the basic assumption of dividing a person into a front-and-back-view image pair. So we can not handle severe self-occlusions. However, this might be solved by combining other representations which are strong on body shape and pose prediction (like the implicit function used in PIFU~\cite{pifu} and the parametric model used in DeepHuman~\cite{deephuman}), and our NormalGAN can act as a detail layer generation module to guarantee fully detailed reconstruction. 
Another direction to tackle this limitation is that using temporal information in RGB-D sequence to recover detailed geometries for current occluded body parts. We believe our method will stimulate future researches on 3D human reconstruction as well as enable more real-scene applications. 

\noindent\textbf{Acknowledgement.} This paper is supported by the National Key Research and Development Program of China [2018YFB2100500] and the NSFC No.61827805 and No.61861166002.

%
%

\begin{thebibliography}{10}
\providecommand{\url}[1]{\texttt{#1}}
\providecommand{\urlprefix}{URL }
\providecommand{\doi}[1]{https://doi.org/#1}

\bibitem{cui2013kinectavatar}
Cui, Y., Chang, W., N{\"o}ll, T., Stricker, D.: Kinectavatar: Fully automatic
  body capture using a single kinect. In: Park, J.I., Kim, J. (eds.) Computer
  Vision - ACCV 2012 Workshops. pp. 133--147. SPRINGER Berlin Heidelberg,
  Berlin, Heidelberg (2013)

\bibitem{Dou_Fusion4D_2017}
Dou, M., Khamis, S., Degtyarev, Y., Davidson, P., Fanello, S.R., Kowdle, A.,
  Escolano, S.O., Rhemann, C., Kim, D., Taylor, J., et~al.: Fusion4d: Real-time
  performance capture of challenging scenes. ACM Trans. Graph.  \textbf{35}(4)
  (Jul 2016). \doi{10.1145/2897824.2925969},
  \url{https://doi.org/10.1145/2897824.2925969}

\bibitem{fankhauser2015kinect}
Fankhauser, P., Bloesch, M., Rodriguez, D., Kaestner, R., Hutter, M., Siegwart,
  R.: Kinect v2 for mobile robot navigation: Evaluation and modeling. In: 2015
  International Conference on Advanced Robotics (ICAR). pp. 388--394. IEEE
  (2015)

\bibitem{moulding}
Gabeur, V., Franco, J., Martin, X., Schmid, C., Rogez, G.: Moulding humans:
  Non-parametric 3d human shape estimation from single images. CoRR
  \textbf{abs/1908.00439} (2019), \url{http://arxiv.org/abs/1908.00439}

\bibitem{Guan:ICCV:2009}
Guan, P., Weiss, A., Balan, A., Black, M.J.: Estimating human shape and pose
  from a single image. In: Int. Conf. on Computer Vision, ICCV. pp. 1381--1388
  (2009)

\bibitem{GuoTOG2017}
Guo, K., Xu, F., Yu, T., Liu, X., Dai, Q., Liu, Y.: Real-time geometry, albedo,
  and motion reconstruction using a single rgb-d camera. ACM Trans. Graph.
  \textbf{36}(3),  32:1--32:13 (Jun 2017). \doi{10.1145/3083722},
  \url{http://doi.acm.org/10.1145/3083722}

\bibitem{han2013high}
Han, Y., Lee, J., Kweon, I.S.: High quality shape from a single rgb-d image
  under uncalibrated natural illumination. International Conference on Computer
  Vision pp. 1617--1624 (2013)

\bibitem{innmann2016volume}
Innmann, M., Zollh{\"o}fer, M., Nie{\ss}ner, M., Theobalt, C., Stamminger, M.:
  Volumedeform: Real-time volumetric non-rigid reconstruction. In: European
  Conference on Computer Vision (ECCV). vol.~9912, pp. 362--379. SPRINGER,
  Amsterdam (2016)

\bibitem{Izadi:2011}
Izadi, S., Kim, D., Hilliges, O., Molyneaux, D., Newcombe, R., Kohli, P.,
  Shotton, J., Hodges, S., Freeman, D., Davison, A., Fitzgibbon, A.:
  Kinectfusion: Real-time 3d reconstruction and interaction using a moving
  depth camera. In: Proc. UIST. pp. 559--568. ACM (2011)

\bibitem{johnson2016perceptual}
Johnson, J., Alahi, A., Feifei, L.: Perceptual losses for real-time style
  transfer and super-resolution. Computer Vision and Pattern Recognition
  (2016)

\bibitem{kanazawa2018end-to-end}
Kanazawa, A., Black, M.J., Jacobs, D.W., Malik, J.: End-to-end recovery of
  human shape and pose. Computer Vision and Pattern Recognition pp. 7122--7131
  (2018)

\bibitem{chao2018ArticulatedFusion}
Li, C., Zhang, Z., Guo, X.: Articulatedfusion: Real-time reconstruction of
  motion, geometry and segmentation using a single depth camera. In: European
  Conference on Computer Vision (ECCV). pp. 324--40. SPRINGER, Munich (2018)

\bibitem{SMPL:2015}
Loper, M., Mahmood, N., Romero, J., Pons-Moll, G., Black, M.J.: {SMPL}: A
  skinned multi-person linear model. ACM Trans. Graphics (Proc. SIGGRAPH Asia)
  \textbf{34}(6),  248:1--248:16 (Oct 2015)

\bibitem{natsume2019siclope:}
Natsume, R., Saito, S., Huang, Z., Chen, W., Ma, C., Li, H., Morishima, S.:
  Siclope: Silhouette-based clothed people. Computer Vision and Pattern
  Recognition pp. 4480--4490 (2019)

\bibitem{newcombe2015dynamic}
Newcombe, R.A., Fox, D., Seitz, S.M.: Dynamicfusion: Reconstruction and
  tracking of non-rigid scenes in real-time. In: IEEE Conference on Computer
  Vision and Pattern Recognition (CVPR). pp. 343--352. IEEE, Boston (2015)

\bibitem{SMPL-X:2019}
Pavlakos, G., Choutas, V., Ghorbani, N., Bolkart, T., Osman, A.A.A., Tzionas,
  D., Black, M.J.: Expressive body capture: 3d hands, face, and body from a
  single image. In: Proceedings IEEE Conf. on Computer Vision and Pattern
  Recognition (CVPR) (Jun 2019), \url{http://smpl-x.is.tue.mpg.de}

\bibitem{pavlakos2018learning}
Pavlakos, G., Zhu, L., Zhou, X., Daniilidis, K.: Learning to estimate 3d human
  pose and shape from a single color image. Computer Vision and Pattern
  Recognition pp. 459--468 (2018)

\bibitem{petrov1993on}
Petrov, A.P.: On obtaining shape from color shading. Color Research and
  Application  \textbf{18}(6),  375--379 (1993)

\bibitem{rhodin2016general}
Rhodin, H., Robertini, N., Casas, D., Richardt, C., Seidel, H., Theobalt, C.:
  General automatic human shape and motion capture using volumetric contour
  cues. Computer Vision and Pattern Recognition  (2016)

\bibitem{ronneberger2015u-net:}
Ronneberger, O., Fischer, P., Brox, T.: U-net: Convolutional networks for
  biomedical image segmentation. Computer Vision and Pattern Recognition
  (2015)

\bibitem{pifu}
Saito, S., Huang, Z., Natsume, R., Morishima, S., Kanazawa, A., Li, H.: Pifu:
  Pixel-aligned implicit function for high-resolution clothed human
  digitization. CoRR  \textbf{abs/1905.05172} (2019),
  \url{http://arxiv.org/abs/1905.05172}

\bibitem{simonyan2014very}
Simonyan, K., Zisserman, A.: Very deep convolutional networks for large-scale
  image recognition. Computer Vision and Pattern Recognition  (2014)

\bibitem{slavcheva2017cvpr}
Slavcheva, M., Baust, M., Cremers, D., Ilic, S.: {KillingFusion: Non-rigid 3D
  Reconstruction without Correspondences}. In: IEEE Conference on Computer
  Vision and Pattern Recognition (CVPR). pp. 5474--5483. IEEE, Honolulu (2017)

\bibitem{Slavcheva_2018_CVPR}
Slavcheva, M., Baust, M., Ilic, S.: Sobolevfusion: 3d reconstruction of scenes
  undergoing free non-rigid motion. In: The IEEE Conference on Computer Vision
  and Pattern Recognition (CVPR). pp. 2646--2655. IEEE, Salt Lake City (June
  2018)

\bibitem{smith2019facsimile}
Smith, D., Loper, M., Hu, X., Mavroidis, P., Romero, J.: Facsimile: Fast and
  accurate scans from an image in less than a second. In: The IEEE
  International Conference on Computer Vision (ICCV). pp. 5330--5339 (2019)

\bibitem{sterzentsenko2019self-supervised}
Sterzentsenko, V., Saroglou, L., Chatzitofis, A., Thermos, S., Zioulis, N.,
  Doumanoglou, A., Zarpalas, D., Daras, P.: Self-supervised deep depth
  denoising. arXiv: Computer Vision and Pattern Recognition  (2019)

\bibitem{tong2012scanning}
Tong, J., Zhou, J., Liu, L., Pan, Z., Yan, H.: Scanning 3d full human bodies
  using kinects. IEEE Transactions on Visualization and Computer Graphics
  \textbf{18}(4),  643--650 (2012)

\bibitem{varol2018bodynet:}
Varol, G., Ceylan, D., Russell, B.C., Yang, J., Yumer, E., Laptev, I., Schmid,
  C.: Bodynet: Volumetric inference of 3d human body shapes. Computer Vision
  and Pattern Recognition  (2018)

\bibitem{wang2018high}
Wang, T.C., Liu, M.Y., Zhu, J.Y., Tao, A., Kautz, J., Catanzaro, B.:
  High-resolution image synthesis and semantic manipulation with conditional
  gans. In: Proceedings of the IEEE conference on computer vision and pattern
  recognition. pp. 8798--8807 (2018)

\bibitem{wu2011}
{Wu}, C., {Varanasi}, K., {Liu}, Y., {Seidel}, H., {Theobalt}, C.:
  Shading-based dynamic shape refinement from multi-view video under general
  illumination. In: 2011 International Conference on Computer Vision. pp.
  1108--1115 (Nov 2011). \doi{10.1109/ICCV.2011.6126358}

\bibitem{wu2013}
Wu, C., Stoll, C., Valgaerts, L., Theobalt, C.: On-set performance capture of
  multiple actors with a stereo camera. ACM Transactions on Graphics
  \textbf{32},  1--11 (11 2013). \doi{10.1145/2508363.2508418}

\bibitem{yan2018DDRNet}
Yan, S., Wu, C., Wang, L., Xu, F., An, L., Guo, K., Liu, Y.: Ddrnet: Depth map
  denoising and refinement for consumer depth cameras using cascaded cnns. In:
  ECCV (2018)

\bibitem{yu2013shading-based}
Yu, L., Yeung, S., Tai, Y., Lin, S.: Shading-based shape refinement of rgb-d
  images. Computer Vision and Pattern Recognition pp. 1415--1422 (2013)

\bibitem{yu2017BodyFusion}
Yu, T., Guo, K., Xu, F., Dong, Y., Su, Z., Zhao, J., Li, J., Dai, Q., Liu, Y.:
  Bodyfusion: Real-time capture of human motion and surface geometry using a
  single depth camera. In: IEEE International Conference on Computer Vision
  (ICCV). pp. 910--919. IEEE, Venice (2017)

\bibitem{doublefusion}
Yu, T., Zheng, Z., Guo, K., Zhao, J., Dai, Q., Li, H., Pons{-}Moll, G., Liu,
  Y.: Doublefusion: Real-time capture of human performances with inner body
  shapes from a single depth sensor. In: 2018 {IEEE} Conference on Computer
  Vision and Pattern Recognition, {CVPR} 2018, Salt Lake City, UT, USA, June
  18-22, 2018. pp. 7287--7296 (2018)

\bibitem{Zeng_2013_CVPR}
Zeng, M., Zheng, J., Cheng, X., Liu, X.: Templateless quasi-rigid shape
  modeling with implicit loop-closure. In: The IEEE Conference on Computer
  Vision and Pattern Recognition (CVPR) (June 2013)

\bibitem{zhang1999shape-from-shading:}
Zhang, R., Tsai, P., Cryer, J.E., Shah, M.: Shape-from-shading: a survey. IEEE
  Transactions on Pattern Analysis and Machine Intelligence  \textbf{21}(8),
  690--706 (1999)

\bibitem{deephuman}
Zheng, Z., Yu, T., Wei, Y., Dai, Q., Liu, Y.: Deephuman: 3d human
  reconstruction from a single image. CoRR  \textbf{abs/1903.06473} (2019),
  \url{http://arxiv.org/abs/1903.06473}

\end{thebibliography}

\end{document}